\documentclass[runningheads]{llncs}

\usepackage[utf8]{inputenc} % allow utf-8 input
\usepackage[T1]{fontenc}    % use 8-bit T1 fonts
\usepackage{hyperref}       % hyperlinks
\usepackage{url}            % simple URL typesetting
\usepackage{booktabs}       % professional-quality tables
\usepackage{amsfonts}       % blackboard math symbols
\usepackage{nicefrac}       % compact symbols for 1/2, etc.
\usepackage{microtype}  
\usepackage{graphicx}
\usepackage{multirow}
\usepackage{makecell}
\usepackage{tabularx}
\usepackage{textcomp}
\usepackage{enumitem}
\usepackage{xcolor}
\usepackage{float}
\usepackage[htt]{hyphenat}
\usepackage[square,numbers]{natbib}

\begin{document}
\title{Assessing generalization capability of text ranking models in Polish}
\titlerunning{Assessing generalization capability of text ranking models in Polish}

\author{Sławomir Dadas \and Małgorzata Grębowiec}
\authorrunning{S. Dadas and M. Grębowiec}
\institute{National Information Processing Institute, Warsaw, Poland\\
\email{\{sdadas,mgrebowiec\}@opi.org.pl}}

\maketitle 
\begin{abstract}
Retrieval-augmented generation (RAG) is becoming an increasingly popular technique for integrating internal knowledge bases with large language models. In a typical RAG pipeline, three models are used, responsible for the retrieval, reranking, and generation stages. In this article, we focus on the reranking problem for the Polish language, examining the performance of rerankers and comparing their results with available retrieval models. We conduct a comprehensive evaluation of existing models and those trained by us, utilizing a benchmark of 41 diverse information retrieval tasks for the Polish language. The results of our experiments show that most models struggle with out-of-domain generalization. However, a combination of effective optimization method and a large training dataset allows for building rerankers that are both compact in size and capable of generalization. The best of our models establishes a new state-of-the-art for reranking in the Polish language, outperforming existing models with up to 30 times more parameters.

\end{abstract}

\section{Introduction}
Text information retrieval is one of the most active research areas in natural language processing. The popularity of this field primarily stems from its broad applications and the challenges posed by steadily increasing volumes of unstructured data, whether publicly available on the Internet or processed internally by companies. Information retrieval methods are applied to various problems, including search, question answering, summarization, clustering, or plagiarism detection. Text retrieval originally relied on lexical methods such as BM25 \citep{robertson2009probabilistic}, which, with the advancement of deep learning, began to be replaced or supplemented by neural approaches, particularly those leveraging neural language models \citep{zhao2022dense,zhu2023large}.

\begin{figure}
  \centering
  \includegraphics[scale=0.75]{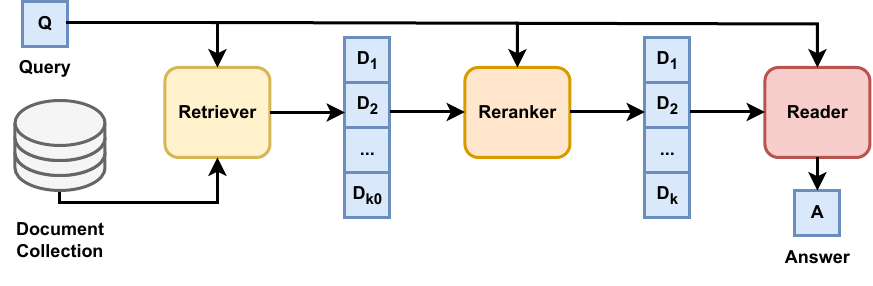}
  \caption{A diagram showing the three-step RAG workflow. In the first step, the retriever model extracts a list of $k0$ best-matching documents from the document collection based on the user's query. Subsequently, the documents are sorted using the reranker, which returns the top $k$ documents ordered by their relevance to the query, where $k \leq k0$. The query is then passed to the reader model, along with the context that includes the text of the retrieved documents. The model is responsible for generating the final answer using the context in either an extractive or abstractive manner.}
  \label{fig:diagram}
\end{figure}

Nowadays, a lot of attention is being paid to building end-to-end systems that are designed to answer a user's question based on data in a local knowledge base. This challenge is typically addressed by integrating multiple cooperating models, each specialized for a specific step in the process. Such approaches are referred to as retrieval-augmented generation (RAG) \citep{lewis2020retrieval,li2022survey}. Figure \ref{fig:diagram} illustrates the typical architecture of the RAG system, consisting of three models: retriever, reranker, and reader. The retriever is responsible for the initial extraction of a set of documents relevant to the user's query. At this stage, the focus is primarily on the model's efficiency - it should be able to quickly select matching documents from a large collection, potentially numbering in the millions or billions. Therefore, lexical methods based on full-text indexes or neural text encoders along with vector indexes are commonly employed. This allows for precomputing representations for documents and storing them in a data structure that enables efficient searching. In the next step, the retrieved documents are sorted using the reranker model. Since the input consists of a small set of pre-selected documents, models used at this stage may offer higher prediction quality at the expense of efficiency. Frequently, employed methods compute similarity for each query-document pair, requiring $n \times m$ comparisons for $n$ queries and $m$ documents per query. In the final step of the process, the query along with the content of the documents is passed to the reader, which generates the final answer. Currently, large language models (LLMs) are most commonly used for this purpose.

This publication focuses on the reranking stage in the RAG process. It is important to note that reranking is an optional step, as it would be possible to skip it and pass the retriever results directly to the reader. The idea of employing additional ranking models is based on the assumption that these models have the ability to better assess the relevance of a document to the query. Therefore, an additional sorting stage should improve the results from the retrieval stage. Research indicates that in the case of supervised learning, in which models are trained and evaluated on a dataset from the same domain, the results of rerankers indeed surpass those obtained by retrievers \citep{yates2021pretrained}. However, in the case of zero-shot evaluation, in which we train a model on a single dataset and evaluate it on multiple datasets with different structures and coming from various domains, this may not necessarily be true. Recent publications on rerankers pay more attention to their generalization ability and zero-shot performance \cite{zhuang2023rankt5,ma2023fine,sun-etal-2023-chatgpt}. The results of these comparisons do not always favor rerankers. For some out-of-domain datasets, these models are outperformed even by baseline methods such as BM25. \citet{rosa2022defense} suggests that generalization capability of text rerankers increases with model size, and only large rerankers with several billion parameters achieve good results in a zero-shot setting. Another aspect worth noting is the significant progress made in the field of dense retrievers over the past years. Methods that were popular a few years ago, such as dense passage retrievers (DPR) \citep{karpukhin-etal-2020-dense}, achieved results worse than BM25 for zero-shot tasks \citep{beir2021,zhang2023}. In contrast, modern dense retrievers like E5 \citep{wang2022text} or FlagEmbeddings \citep{xiao2023c} demonstrate strong performance, surpassing BM25 significantly, as confirmed by benchmarks such as BEIR \citep{beir2021}. During the same period, we have not observed such significant progress for text ranking models, which may suggest that the gap between these two groups of methods has narrowed.

\subsection{Contributions}
The aim of this article is to investigate the performance of rerankers for the Polish language and compare their results with available retrieval models. For this purpose, we use PIRB \citep{pirb2023}, an extensive benchmark covering 41 information retrieval tasks for the Polish language. In this publication, we answer the question of whether the available Polish and multilingual rerankers outperform standalone retrieval-based solutions and whether they demonstrate sufficient generalization capabilities. We also highlight the properties of models and datasets that influence the quality of text reranking, attempting to provide guidelines for when it is worthwhile to include reranker step in the RAG process. More specifically, we make the following contributions:
\begin{itemize}[wide,labelwidth=0pt,labelindent=0pt,itemsep=0pt,topsep=5pt]
\item[$\bullet$]{We evaluate several Polish and multilingual reranking models on a collection of information retrieval tasks for the Polish language. In the evaluation, we consider publicly available models as well as reproduce several baseline methods described in the literature. Our experiments involve rerankers of various scales, ranging from tens of millions to 13 billion parameters.}
\item[$\bullet$]{Using distillation techniques, we train new highly efficient reranking models for the Polish language. Our method involves leveraging the largest available reranker based on the MT5-XXL architecture \citep{xue-etal-2021-mt5} as a teacher model in a knowledge distillation procedure. We employ two distillation methods, one based on directly mimicking the predictions of a large reranker, and the other based on permutation modeling. The best of our models outperforms the teacher model on the PIRB benchmark while being 30 times smaller in terms of the number of parameters.}
\end{itemize}

\subsection{Related work}
With the advancement of deep learning methods, models based on neural networks are finding increasingly more applications in the field of information retrieval, spanning from first-stage retrieval, through reranking, to the extraction of specific information from documents. Early neural reranking methods utilized architectures such as convolutional or recurrent networks \citep{guo2020deep}, leveraging pre-trained word embeddings. However, it was only with the emergence of the transformer architecture that the topic received a significant increase in attention. Today, many high-quality reranker models are available for popular languages such as English and Chinese. Most such solutions are based on encoder architectures like BERT \citep{nogueira2019passage,yates2021pretrained}, or encoder-decoder architectures like T5 \citep{nogueira-etal-2020-document,zhuang2023rankt5}.

Beyond the development of neural methods and architectures, the availability of training data is equally crucial. The progress we observe would not have been possible without the release of the MS MARCO dataset \citep{bajaj2016ms}, comprising approximately 1 million queries and 9 million passages with relevance judgments generated by the users of the Bing search engine. This enabled the training of general-purpose rerankers that handle a wide range of topics and domains. Datasets of a similar scale for languages other than English have only started to emerge recently. Notable among these is mMARCO \citep{bonifacio2021mmarco}, which includes data from the original MS MARCO translated into 13 languages, as well as other recently released large collections such as Mr.TyDi \citep{zhang-etal-2021-mr} and MIRACL \citep{zhang2023}. Using these datasets, several publicly available multilingual rerankers have been trained.

In the case of the Polish language, until recently, the topic of deep learning methods for information retrieval was absent from the scientific literature. The situation began to change only in the last two years, primarily concerning the availability of datasets for information retrieval and question answering. A significant number of individual datasets as well as collections comprising several datasets have been prepared and made available during this period. These include PolQA \citep{rybak2022improving}, MAUPQA \citep{rybak2023maupqa}, BEIR-PL \citep{wojtasik2023beir} (a machine translated version of the English BEIR \citep{beir2021} benchmark), as well as datasets used in the PolEval-2022 Passage Retrieval task \citep{poleval2022}. Most of the mentioned datasets have been combined into the PIRB benchmark \citep{pirb2023}, for which the authors additionally collected 10 new datasets by crawling QA sections on Polish websites. In total, the benchmark included 41 information retrieval tasks for the Polish language. In contrast to data availability, there is still substantial room for improvement when it comes to models. Previous studies have primarily focused on dense retrievers \citep{rybak2023silverretriever,pirb2023}. The topic of reranking has only been addressed by \citet{wojtasik2023beir}. The authors have also provided four reranking models for the Polish language.

\section{Evaluation}
In this chapter, we describe the evaluation results of Polish and multilingual rerankers on the Polish Information Retrieval Benchmark (PIRB). In our experiments, we considered publicly available models as well as models specifically trained by us for this study using publicly available source code. Since reranking is the second stage of the RAG process, to conduct the evaluation, we also used a retrieval model to initially select documents to be sorted by each of the rerankers. For the first-stage retrieval, we utilized {\tt mmlw-retrieval-roberta-large} model, which, at the time of the experiments, was the top-performing retriever in the PIRB benchmark according to the NDCG@10 metric. The model was used to extract the top 100 documents for each query, which were then passed as input to the rerankers. Using the same retriever for all rerankers allowed for a reliable comparison of their results both among themselves and in relation to the performance of the retriever itself. It should be noted that the computational cost of reranking is high, and some datasets in the benchmark contain up to several hundred thousand queries. Considering this, to enable efficient evaluation of multiple models, we limited the number of queries to the first 1000 for each dataset. In the following sections, we describe the evaluated methods and then proceed to the analysis of the results. 

\subsection{Evaluated models}
In our experiments, we included the following publicly available models:
\begin{itemize}[wide,labelwidth=0pt,labelindent=0pt,itemsep=0pt,topsep=5pt]
\item[$\bullet$] \textbf{Polish rerankers} - The first group of methods consists of models trained as a part of experiments conducted by \citet{wojtasik2023beir}. The authors evaluated several baseline retrieval and reranking methods. They trained four rerankers on the training split of the Polish MS MARCO dataset. This includes two types of models and two sizes (base and large) for each type. The first type includes standard cross-encoders with classification head based on the HerBERT language model \citep{mroczkowski-etal-2021-herbert}. The second type utilizes a sequence-to-sequence approach, in which the model takes the query-document pair as input and outputs a "positive" token if the document is relevant to the query or a 
"negative" token otherwise. These rerankers were created by fine-tuning plT5 language model \citep{chrabrowa-etal-2022-evaluation}.
\item[$\bullet$] \textbf{Multilingual rerankers trained on mMARCO} - mMARCO is a multilingual version of the MS MARCO dataset translated to 13 languages by researchers from the University of Campinas, Brazil. After the release of this dataset, the authors also published a set of rerankers trained by them\footnote{\url{https://huggingface.co/unicamp-dl}}. Most of these models are fine-tuned versions of MT5 \citep{xue-etal-2021-mt5} with varying sizes, ranging from a few hundred million to 13 billion parameters. 
\end{itemize}

In addition to the previously provided models, we also trained our own rerankers for Polish, utilizing three popular learning methods from the literature. For each method, we fine-tuned the models in two sizes, leveraging Polish RoBERTa language models \citep{dadas2020pre}. To train the models, we used data from the Polish translation of MS MARCO. We employed the following methods:
\begin{itemize}[wide,labelwidth=0pt,labelindent=0pt,itemsep=0pt,topsep=5pt]
\item[$\bullet$] \textbf{Binary cross entropy (BCE) loss with hard negatives} - The simplest and most popular method for training rerankers is to treat the problem as binary classification, in which relevant query-document pairs are positive samples and non-relevant pairs are negative samples. The effectiveness of this method depends on the selection of appropriate negatives that are difficult to distinguish from positive pairs. In our procedure, we used a set of negatives available in the Sentence-Transformers library\footnote{\url{https://huggingface.co/datasets/sentence-transformers/msmarco-hard-negatives}}. The models were trained for 10 epochs with a batch size of 32 and peak learning rate of 1e-5 with linear decay.
\item[$\bullet$] \textbf{Knowledge distillation with mean squared error (MSE) loss} - Another way of training is to use relevancy scores from another high-quality reranker and teach the new model to mimic those predictions. This is therefore a regression task because we directly use numerical labels denoting the level of relevance between the query and the document. In this procedure, we used publicly available pre-computed set of scores for the MS MARCO dataset\footnote{\url{https://zenodo.org/record/4068216/files/bert_cat_ensemble_msmarcopassage_train_scores_ids.tsv}}. The models were trained for 10 epochs with batch size of 32 and peak learning rate of 1e-5 with linear decay.
\item[$\bullet$] \textbf{Permutation distillation with RankNet loss} - The above two methods are based on simple loss functions that consider each query-document pair independently. More sophisticated techniques for training rerankers propose listwise approaches, in which the loss is calculated based on the query and the set of all documents assigned to that query. One such method is RankNet \citep{burges2005learning}, in which loss is computed based on the relative order of documents sorted by their relevance to the query. The technique has been applied recently by \citet{sun-etal-2023-chatgpt}, who proposed to generate sorted lists of relevant documents by prompting large language models, and then train smaller neural rerankers on this data. We applied the same idea to train our models. First, we generated 10,000 samples from the MS MARCO dataset, where each sample consisted of a query and a sorted list of 20 relevant documents. Then we used the scripts provided by the authors to fine-tune the models for Polish. For training, we employed the hyperparameters suggested in their article \citep{sun-etal-2023-chatgpt} - the models were trained for two epochs with a batch size of 32 and a constant learning rate of 5e-5.
\end{itemize}

\subsection{Evaluation results}
The results of our experiments are shown in Table \ref{tab:eval_baselines}. The table is divided into sections based on the size of the evaluated models. The first group consists of models of small and base sizes, ranging from tens to a maximum of a few hundred million parameters. The subsequent groups include large models (>300 million parameters) and the largest models (>3 billion). For comparison, the table also includes the results of first-stage retrieval methods: BM25 and {\tt mmlw-retrieval-roberta-large} dense retriever. The latter model is particularly important since it was used to generate candidate documents for reranking. Therefore, we expect rerankers to achieve results higher than those obtained by this model, since only improving on the first-stage results makes the reranker useful. Scores higher than the retriever are shown in blue.

The most important observation is that the majority of rerankers were unable to surpass the results of the retrieval stage. Only two models achieved a higher average NDCG@10 metric value for the entire benchmark – the largest from the evaluated models ({\tt mt5-13b-mmarco-100k}) and the {\tt polish-roberta-large-v2} model fine-tuned using the RankNet method. The situation looks better when broken down into individual task groups. Some types of datasets appear to be easier for rerankers (e.g., PolEval or MAUPQA), while others pose significant challenges (e.g., Web Datasets or BEIR-PL). Since the data from the PolEval and MAUPQA groups were prepared by the same authors, we can speculate that some steps in the process of building these datasets, such as the initial selection of documents for labeling, might favor reranking models. On the other hand, Web Datasets and BEIR-PL are characterized by considerable data diversity, as they represent compilations of datasets from multiple sources or prepared by different authors, sharing few common features.

\begin{table}
  \centering
  \caption{Evaluation results on PIRB. We report the average NDCG@10 metric values for all datasets (41 tasks) and for the individual task groups. Rerankers are divided into groups based on the number of parameters. The highest NDCG@10 value in each group is underlined. Highlighted in blue are the results of rerankers that are higher than those achieved first-stage retrieval. $*$ indicates models trained by us.}
  \aboverulesep=0ex
  \belowrulesep=0ex
  \setlength{\tabcolsep}{1pt}
  \renewcommand{\arraystretch}{0.8}
  \begin{tabular}{l|c|ccccc}
    \toprule
    \textbf{Model name} & \scalebox{0.9}[1.0]{Average} & \scalebox{0.9}[1.0]{PolEval} & \scalebox{0.9}[1.0]{WebDS} & \scalebox{0.9}[1.0]{BEIR} & \scalebox{0.9}[1.0]{MAUPQA} & \scalebox{0.9}[1.0]{Other} \\
    \hline
    \multicolumn{7}{l}{\textbf{Retrieval only}}\\
    \hline
    \makecell[l]{\scalebox{0.9}[1.0]{BM25}} & 41.85 & 45.51 & 47.27 & 33.26 & 38.64 & 71.09 \\
    \makecell[l]{\scalebox{0.9}[1.0]{sdadas/mmlw-retrieval-roberta-large}} & \underline{58.53} & \underline{62.72} & \underline{67.43} & \underline{53.19} & \underline{50.08} & \underline{83.85} \\
    \hline
    \multicolumn{7}{l}{\textbf{Small \& base models}}\\
    \hline
    \makecell[l]{\scalebox{0.9}[1.0]{unicamp-dl/mMiniLM-L6-v2-mmarco-v2}} & 49.28 & 58.76 & 45.41 & 40.61 & 49.80 & 78.10 \\
    \makecell[l]{\scalebox{0.9}[1.0]{clarin-knext/herbert-base-reranker-msmarco}} & 49.77 & 59.87 & 46.52 & 41.47 & 49.22 & 78.12 \\
    \makecell[l]{\scalebox{0.9}[1.0]{nreimers/mmarco-mMiniLMv2-L12-H384-v1}} & 53.38 & \color{blue} 64.69 & 51.02 & 44.37 & \color{blue} \underline{52.37} & 80.12 \\
    \makecell[l]{\scalebox{0.9}[1.0]{unicamp-dl/mt5-base-mmarco-v2}} & 53.98 & \color{blue} 63.29 & 56.55 & 44.31 & \color{blue} 50.84 & 81.72 \\
    \makecell[l]{\scalebox{0.9}[1.0]{clarin-knext/plt5-base-msmarco}} & \underline{55.39} & \color{blue} 63.40 & \underline{60.65} & 45.69 & \color{blue} 51.29 & 81.61 \\
    \makecell[l]{\scalebox{0.9}[1.0]{polish-roberta-base-v2 (BCE)*}} & 52.56 & \color{blue} 64.23 & 51.35 & 44.27 & 49.73 & 79.67 \\
    \makecell[l]{\scalebox{0.9}[1.0]{polish-roberta-base-v2 (RankNet)*}} & 52.75 & \color{blue} \underline{65.36} & 53.78 & 42.38 & 49.27 & \underline{81.90} \\
    \makecell[l]{\scalebox{0.9}[1.0]{polish-roberta-base-v2 (MSE)*}} & 54.12 & \color{blue} 64.57 & 53.51 & \underline{46.34} & \color{blue} 50.99 & 81.81 \\
    \hline
    \multicolumn{7}{l}{\textbf{Large models (> 300m params)}} \\
    \hline
    \makecell[l]{\scalebox{0.9}[1.0]{clarin-knext/herbert-large-msmarco}} & 52.96 & \color{blue} 66.22 & 51.19 & 42.23 & \color{blue} 51.94 & 79.64 \\
    \makecell[l]{\scalebox{0.9}[1.0]{clarin-knext/plt5-large-msmarco}} & 57.36 & \color{blue} 65.97 & \underline{64.11} & 46.98 & \color{blue} 52.25 & \color{blue} 84.47 \\
    \makecell[l]{\scalebox{0.9}[1.0]{polish-roberta-large-v2 (BCE)*}} & 56.83 & \color{blue} 69.04 & 60.29 & 45.80 & \color{blue} \underline{52.64} & \color{blue} 84.29 \\
    \makecell[l]{\scalebox{0.9}[1.0]{polish-roberta-large-v2 (MSE)*}} & 57.28 & \color{blue} 68.02 & 60.62 & \underline{48.00} & \color{blue} 52.56 & \color{blue} 83.98 \\
    \makecell[l]{\scalebox{0.9}[1.0]{polish-roberta-large-v2 (RankNet)*}} & \color{blue} \underline{58.57} & \color{blue} \underline{71.70} & 63.91 & 47.54 & \color{blue} 52.36 & \color{blue} \underline{86.42} \\
    \hline
    \multicolumn{7}{l}{\textbf{XL \& XXL models (> 3b params)}} \\
    \hline
    \makecell[l]{\scalebox{0.9}[1.0]{unicamp-dl/mt5-3b-mmarco-100k-kdd-alltrain}} & 56.97 & \color{blue} 65.96 & 64.34 & 46.37 & \color{blue} 51.10 & \color{blue} 85.96 \\
    \makecell[l]{\scalebox{0.9}[1.0]{unicamp-dl/mt5-13b-mmarco-100k}} & \color{blue} \underline{61.63} & \color{blue} \underline{70.54} & \color{blue} \underline{72.13} & \underline{51.62} & \color{blue} \underline{53.62} & \color{blue} \underline{86.25} \\
    \bottomrule
  \end{tabular}
  \label{tab:eval_baselines}
\end{table}

\begin{figure}
  \centering
  \includegraphics[scale=0.50]{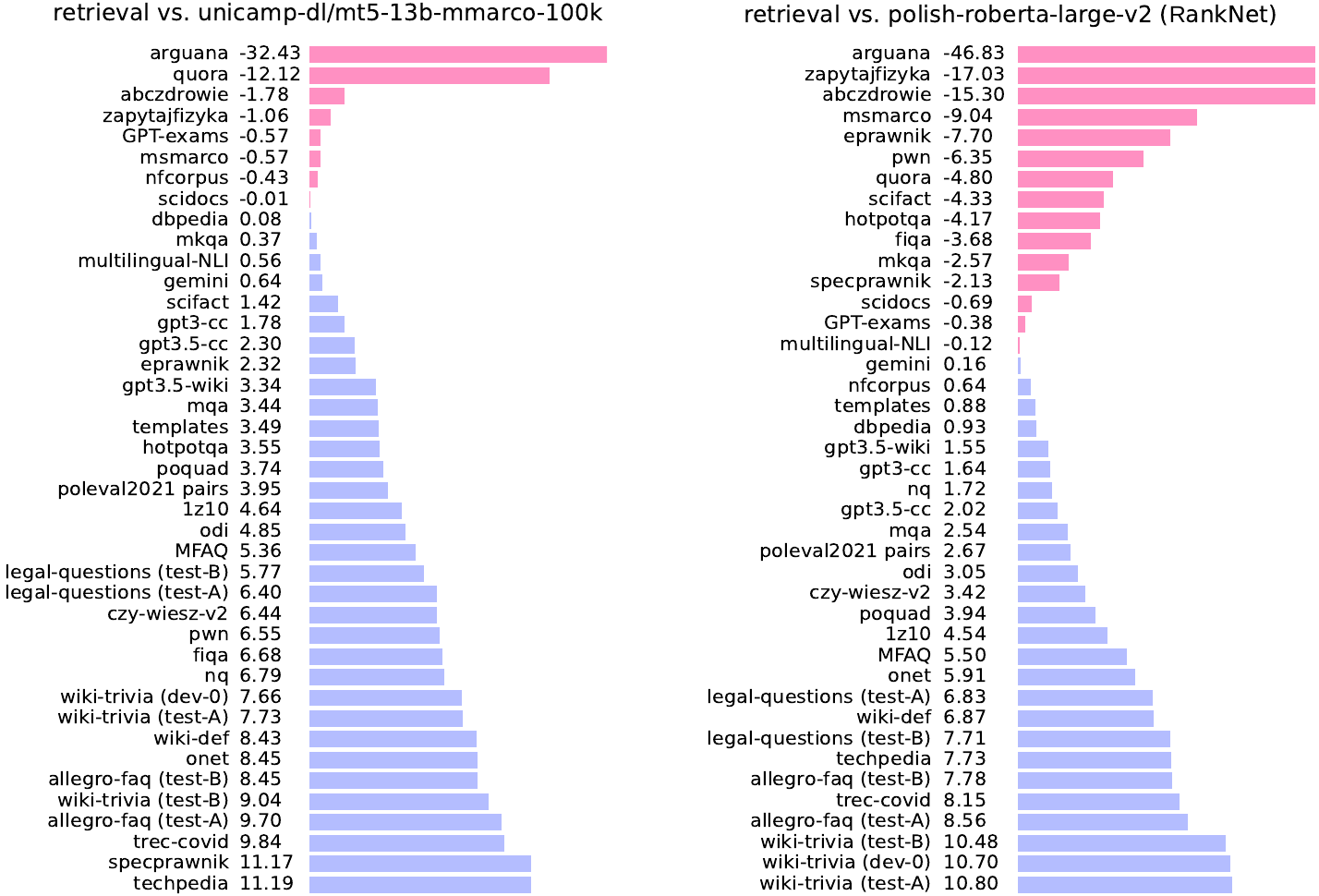}
  \caption{Comparison of NDCG@10 scores between first-stage retrieval and reranking for the top two rerankers, broken down into individual datasets from the PIRB benchmark.}
  \label{fig:comparison}
\end{figure}

In order to further investigate the underlying factors affecting the performance of the rerankers, we compared their results against those achieved by the retriever, broken down into individual datasets. Charts illustrating this comparison for two best-performing rerankers are shown in Figure \ref{fig:comparison}. Each chart shows 41 datasets from PIRB sorted by the absolute improvement in NDCG@10 metric values compared to the retrieval stage.

Particularly noteworthy are the \emph{arguana} and \emph{quora} datasets, as both rerankers perform poorly on them. For the 13B model, these are the only two datasets where this model significantly lags behind the retriever. For the rest, the differences are either small, or the reranker demonstrates a clear advantage. This can be explained by the fact that these two datasets are not typical examples of question answering tasks. In the case of the \emph{arguana}, the goal is to find the best matching counterarguments to a given argument, while the \emph{quora} task involves finding duplicates for a given question. It is understandable that models fine-tuned on QA-oriented data may not generalize well to other types of information retrieval tasks. However, it is worth noting that dense retrievers handle such problems better, even when not trained on similar data.

The smaller of the two rerankers performs worse on a larger number of datasets. We can observe that these are often specialized datasets related to a narrow domain of knowledge. Examples include \emph{zapytajfizyka} (physics), \emph{abczdrowie} (medicine), \emph{eprawnik} (law), \emph{pwn} (linguistics), \emph{scifact} (science), \emph{fiqa} (finance), \emph{specprawnik} (law). In the case of the larger reranker, there is also a slight decrease in quality on some of these datasets, but overall, this model performs significantly better on specialized domains. This confirms the observation of \citet{rosa2022defense} that with an increase in the number of model parameters, its out-of-domain performance also improves. In summary, most of the existing Polish and multilingual rerankers fall short in quality compared to modern dense retrievers, and there is no reason to use them in a zero-shot setting. The solution is either fine-tuning a smaller model to the specific domain or task, or using rerankers with a larger number of parameters that possess better generalization capabilities.

\section{Knowledge distillation}
In the previous chapter, we demonstrated that of the available Polish and multilingual rerankers, only the largest MT5-XXL model performed significantly better than the dense retriever while demonstrating a reasonably good generalization ability for out-of-distribution data. Unfortunately, such large rerankers have limited practical applications due to their high computational requirements and long query processing time. With this in mind, in a subsequent experiment we verified whether it was possible to train smaller models using knowledge distillation methods to achieve performance comparable to the 13B model. In this section, we summarize our findings.

We applied MSE and RankNet methods described in the previous section, utilizing the same language models, i.e. \texttt{polish-roberta-base-v2} and \texttt{polish-roberta-large-v2}. The main difference was using a much larger dataset and employing the 13B model as a teacher instead of publicly available pre-computed set of predictions. Previously, a subset of the MS MARCO training data was used. In this experiment, we expanded the training dataset to include the following parts: 1) All queries from the Polish MS MARCO training split (800k queries); 2) The ELI5\footnote{\url{https://huggingface.co/datasets/eli5}} dataset translated to Polish (over 500k queries); and 3) A collection of questions and answers crawled from the Polish website \emph{znanylekarz} (approximately 100k queries). In total, the compiled dataset consisted of 1.4 million queries and 10 million documents. Then, for each query, we retrieved a set of best-matching documents using three retrieval methods with different characteristics: BM25, dense retriever, and SPLADE \citep{formal2022distillation}. Each method returned the top 16 results, so after removing duplicates we obtained at most 48 documents per query. In the last step, we used the 13B model to score all query-document pairs. For the MSE method, the scores were used directly for training. In the case of the RankNet method, for each query, we created an ordered list by sorting the documents according to their score. We used the same training hyperparameters as in the previous experiment, except for linear decay and a lower learning rate of 2e-5 for the RankNet method.

The results of the newly trained models are shown in Table \ref{tab:eval_distil}. We can observe that the MSE distillation yielded reasonably good results, with both models performing better than previously evaluated methods with a similar number of parameters. However, RankNet distillation proved to be even more effective. The large model trained with this method achieved performance higher than the teacher model, while the base model also demonstrated strong results, outperforming both MSE rerankers. The best of our models achieved the average NDCG@10 of 62.65 on the PIRB benchmark, over 1 point higher than the teacher. It also has 30 times fewer parameters (435M vs. 13B) and is over 33 times faster. The average number of queries per second during the evaluation on a single Nvidia A6000 GPU was 0.15 for the teacher model, 5.0 for \texttt{polish-roberta-large-v2} and 8.4 for \texttt{polish-roberta-base-v2}.

\begin{table}
  \centering
  \caption{Results of the distilled models on the PIRB benchmark. We report the average NDCG@10 metric values for all datasets and for the individual task groups.}
  \aboverulesep=0ex
  \belowrulesep=0ex
  \renewcommand{\arraystretch}{0.9}
  \begin{tabular}{l|c|ccccc}
    \toprule
    \textbf{Model name} & \scalebox{0.9}[1.0]{Average} & \scalebox{0.9}[1.0]{PolEval} & \scalebox{0.9}[1.0]{WebDS} & \scalebox{0.9}[1.0]{BEIR} & \scalebox{0.9}[1.0]{MAUPQA} & \scalebox{0.9}[1.0]{Other} \\
    \hline
    \multicolumn{7}{l}{\textbf{Retrieval only}}\\
    \hline
    \makecell[l]{\scalebox{0.9}[1.0]{sdadas/mmlw-retrieval-roberta-large}} & 58.53 & 62.72 & 67.43 & 53.19 & 50.08 & 83.85 \\
    \hline
    \multicolumn{7}{l}{\textbf{Rerankers}}\\
    \hline
    \makecell[l]{\scalebox{0.9}[1.0]{polish-roberta-base-v2 (MSE)}} & 57.50 & \color{blue} 64.78 & 66.95 & 48.08 & \color{blue} 50.68 & 82.15 \\
    \makecell[l]{\scalebox{0.9}[1.0]{polish-roberta-large-v2 (MSE)}} & \color{blue} 60.27 & \color{blue} 67.54 & \color{blue} 71.44 & 51.13 & \color{blue} 51.92 & \color{blue} 84.87 \\
    \makecell[l]{\scalebox{0.9}[1.0]{polish-roberta-base-v2 (RankNet)}} & \color{blue} 60.32 & \color{blue} 67.92 & \color{blue} 70.39 & 51.61 & \color{blue} 52.48 & 83.36 \\
    \makecell[l]{\scalebox{0.9}[1.0]{polish-roberta-large-v2 (RankNet)}} & \color{blue} \underline{62.65} & \color{blue} \underline{70.77} & \color{blue} \underline{73.81} & \color{blue} \underline{54.05} & \color{blue} 53.54 & \color{blue} 86.01 \\
    \makecell[l]{\scalebox{0.9}[1.0]{unicamp-dl/mt5-13b-mmarco-100k (teacher)}} & \color{blue} 61.63 & \color{blue} \color{blue} 70.54 & \color{blue} 72.13 & 51.62 & \color{blue} \underline{53.62} & \color{blue} \underline{86.25} \\
    \bottomrule
  \end{tabular}
  \label{tab:eval_distil}
\end{table}

The experiment shows that by using a sufficiently large training dataset, we were able to match and even surpass the performance of the multi-billion teacher model. It also demonstrates clear advantage of permutation distillation over the traditional distillation method based on mean squared error loss. In \citet{sun-etal-2023-chatgpt} study for English, they trained models using this technique on datasets of up to 100,000 queries. We believe that combining this method with larger and more diverse training datasets would improve performance for English and multilingual models as well.

\section{Conclusions}
In this publication, we have evaluated the performance of Polish and multilingual text ranking models on PIRB, an exhaustive benchmark covering 41 information retrieval tasks for Polish. Most of the rerankers we have analyzed perform worse than strong dense retriever for Polish, and their generalization ability is low, except for models with a large number of parameters. In our experiments, we also trained a set of new models, including baseline methods as well as high-quality rerankers distilled from the largest available multilingual model based on the MT5-XXL architecture. We showed that with a diverse dataset and an effective training method, it is possible to build a compact reranker that achieves results competitive with significantly larger models. The best of our models sets new state-of-the-art for Polish reranking tasks, outperforming existing solutions.

\bibliographystyle{splncsnat}
\bibliography{references}

\end{document}